\documentclass[journal]{IEEEtran}
\usepackage{url}
\usepackage{cite}
\usepackage{array}
\usepackage{stfloats}
\usepackage{graphicx}
\usepackage{booktabs}
\usepackage{makecell}
\usepackage{mathrsfs}
\usepackage{multirow}
\usepackage[table]{xcolor}
\usepackage{threeparttable}
\usepackage[caption=false]{subfig}
\usepackage{amsmath,amsfonts,amssymb}
\usepackage[pagebackref,breaklinks,colorlinks,hyperfootnotes=true]{hyperref}
\usepackage[capitalize]{cleveref}

\crefname{section}{Sec.}{Secs.}
\crefname{table}{Tab.}{Tabs.}
\Crefname{equation}{Eq.}{Eqs.}
\crefname{figure}{Fig.}{Figs.}
\def\eg{\emph{e.g}.} 
\def\ie{\emph{i.e}.} 
\def\cf{\emph{cf}.} \def\wrt{w.r.t.}
\def\etc{\emph{etc}.} \def\etal{\emph{et al}.}

\begin{document}
\title{Threatening Patch Attacks on Object Detection\\in Optical Remote Sensing Images
}

\author{Xuxiang Sun,
	Gong Cheng,~\IEEEmembership{Member,~IEEE,}
	Lei Pei,
	Hongda Li,
	and Junwei Han,~\IEEEmembership{Fellow,~IEEE}
\thanks{This work was supported in part by the National Natural Science Foundation of China under Grant 62136007, in part by the Natural Science Basic Research Program of Shaanxi under Grants 2021JC-16 and 2023-JC-ZD-36, in part by the Fundamental Research Funds for the Central Universities, in part by the Guangdong Basic and Applied Basic Research Foundation under Grant 2021B1515020072, and in part by the Innovation Foundation for Doctor Dissertation of Northwestern Polytechnical University under Grant CX2022054. (Gong Cheng is the corresponding author).

Xuxiang Sun, Gong Cheng, Lei Pei, and Hongda Li are with the School of Automation, Northwestern Polytechnical University, Xi’an 710129, China, and also with the Research and Development Institute of Northwestern Polytechnical University in Shenzhen, Shenzhen 518057, China. (e-mail: gcheng@nwpu.edu.cn).

Junwei Han is with the School of Automation, Northwestern Polytechnical University, Xi’an 710129, China.

Digital Object Identifier xx.xxxx/TGRS.xxxx.xxxxxx
}}

\markboth{IEEE TRANSACTIONS ON GEOSCIENCE AND REMOTE SENSING}
{Shell \MakeLowercase{\textit{et al.}}: A Sample Article Using IEEEtran.cls for IEEE Journals}

\maketitle
\begin{abstract}
Advanced Patch Attacks~(PAs)~on object detection in natural images have pointed out the great safety vulnerability in methods based on deep neural networks. However, little attention has been paid to this topic in Optical Remote Sensing Images (O-RSIs). To this end, we focus on this research,~\ie, PAs on object detection in O-RSIs, and propose a more Threatening PA without the scarification of the visual quality, dubbed TPA. Specifically, to address the problem of inconsistency between~local and global landscapes in existing patch selection schemes, we propose leveraging the First-Order Difference (FOD) of the~objective function before and after masking to select the sub-patches to be attacked. Further, considering the problem of gradient inundation when applying existing coordinate-based loss to PAs directly, we design an IoU-based objective function specific for PAs, dubbed Bounding box Drifting Loss (BDL), which pushes the detected bounding boxes far from the initial ones until there are no intersections between them. Finally, on two widely used benchmarks,~\ie, DIOR and DOTA, comprehensive evaluations of our TPA with four typical detectors (Faster R-CNN, FCOS, RetinaNet, and YOLO-v4) witness its remarkable effectiveness. To the best of our knowledge, this is the first attempt to study the PAs on object detection in O-RSIs, and we hope this work can get our readers interested in studying this topic.
\end{abstract}

\section{Introduction}\label{sec1}
\IEEEPARstart{D}{raw} on the powerful representation ability of Deep Neural Networks~(DNNs), a great deal of revolutionary achievements have been made in the aspect of image understanding technology~\cite{szegedy2016rethinking, szegedy2017inception, ren2015faster, tian2019fcos, lin2017focal, bochkovskiy2020yolov4, redmon2016you, he2016deep, xie2017aggregated, cheng2022class}. Similarly, the technology of understanding Optical Remote Sensing Images~(O-RSIs)~has made great progress~\cite{cheng2021anchor, xia2018dota, li2022semi, niu2022multi, pei2021multi, li2023instance, yao2022improving}. Nevertheless, the exposed adversarial vulnerability~\cite{goodfellow2014explaining, szegedy2013intriguing} of DNN leaves great security concerns, hindering their widespread applications.

Facing this security hazard, many researchers devote themselves to studying adversarial robustness  \cite{goodfellow2014explaining, szegedy2013intriguing, carlini2017towards, madry2017towards, moosavi2016deepfool, tramer2018ensemble}. Recently, the security concerns of DNN-based deep learning methods in O-RSIs have also received progressive attention~\cite{cheng2021perturbation, xu2022ai, czaja2018adversarial, xu2020assessing}. Among them, little attention has been paid to the adversarial vulnerability in O-RSI object detection, an essential and typical research field in O-RSI understanding.
\begin{figure}[t]
	\centering
	\includegraphics[width=0.95\linewidth]{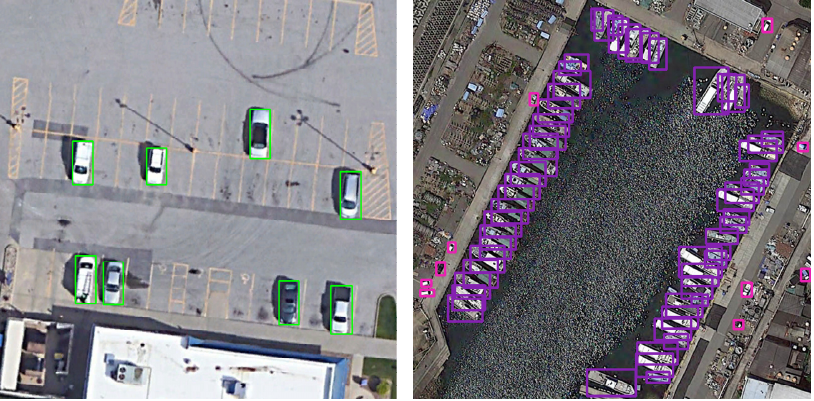}
	\caption{\textbf{Visualization of samples drawn from DIOR and DOTA.}~Here we can see that the distribution of objects in remote sensing images are globally sparse and locally dense.}\label{fig1}
\end{figure}

To date, there are three widely studied attack schemes for object detection,~\ie, the Full-Scale Attacks~(FSAs)~\cite{xie2017adversarial, chen2019shapeshifter, lirobust, zhang2020contextual, nezami2021pick}, the Patch Attacks~(PAs)~\cite{wu2020dpattack, zhao2020object, huang2021rpattack}, and the Adversarial Patches~(APs)~\cite{brown2017adversarial, liu2018dpatch, lee2019physical}. In short, FSAs perturb the whole~image, and only the pixels in some specific regions are perturbed in PAs. Compared to FSAs and PAs, the adversarial examples generated by APs are more human-perceptible, and all the~targets in an image share the same pattern. The only difference among these APs is their physical parameters (location, angle, scale,~\etc). Given the intrinsic property of O-RSIs,~\ie, the~objects in O-RSIs are characterized by globally sparse and locally dense ( \cf, \cref{fig1} for better visualization), PAs exhibit more threats than the other attacks for O-RSIs, on account of their region-efficiency and visual-imperceptibility.

In general, the attack scheme of PAs consists of two critical steps,~\ie, the patch selection scheme and the objective function. For the former, recent research~\cite{huang2021rpattack} proposes to leverage the norm of the gradients passed from the objective function to select the most critical sub-patches. However, adding perturbation is to produce a significant function drop, but the gradient is defined within a tiny neighborhood of data points. Thus, it could not simulate the masking manipulation in PAs. Besides, previous study~\cite{tramer2018ensemble}~has shown that the direction of gradients does not always align with the optimal direction (\cf~\cref{fig2} for more details). That is, for high-dimensional nonlinear functions such as DNNs, the local landscape and the global landscape around a data point are usually inconsistent. To this light, we propose a patch selection scheme based on First-Order Difference~(FOD), which first calculates the FOD of the objective function by masking the sub-patches and selects top-k sub-patches with the largest FOD. In this way, we could find the most critical sub-patches within a relatively larger neighborhood.

Besides, the objective functions leveraged in existing PAs~\cite{huang2021rpattack, zhao2020object, wu2020dpattack}~only focus on the classification branch without the attack on the Bounding box~(Bbox)~regression. Fortunately, a commonly adopted loss~\cite{lirobust}~in the field of FSAs designs the Coordinate-Based Loss~(CBL)~to make all the Bboxes cover the entire image. However, the perturbed regions in PAs are not large enough to force the detected Bboxes to cover the whole image, especially for the images with the characteristic of global sparse such as O-RSIs~(\cf~\cref{fig1}). Then, the gradient of CBL will exist over the entire attack progress with a larger magnitude than that passed from the classification branch~\cite{lirobust}. Consequently, the gradients passed from the classification head may be at risk of being inundated by those passed from the regression branch. To this end, we propose a Bbox Drifting Loss~(BDL) to merely reduce the Intersection over Union (IoU) between the detected Bboxes and the initial ones so as to avoid the problem of gradient inundation when applying CBL to PAs directly.

Finally, we validate the effectiveness of our method on DIOR~\cite{cheng2021anchor} and DOTA~\cite{xia2018dota}, respectively. Here, a total of seven typical detectors are utilized to evaluate the general effectiveness of our method. Specifically, four kinds of victim detectors including Faster R-CNN~\cite{ren2015faster}, RetinaNet~\cite{lin2017focal}, FCOS~\cite{tian2019fcos}, and Yolo-v4~\cite{bochkovskiy2020yolov4} are leveraged for the evaluations, and we equip the first three detectors with two backbones,~\ie, ResNet-50~\cite{he2016deep} with Feature Pyramid Networks (FPN)~\cite{lin2017feature} and ResNet-101~\cite{he2016deep} with FPN~\cite{lin2017feature}. Throughout the comprehensive evaluations, our TPA achieves the most threatening results.

In summary, our contributions are:
\begin{itemize}
	\item The threats of patch attacks on object detection in O-RSIs are exhibited for the first time in this paper, which provides the preliminary empirical evidence for the safety concern when applying DNN-based methods to practical deployment.
	\item We propose FOD patch selection scheme to boost the visual-efficiency of patch attacks. It imitates the attack scheme in PAs by masking the sub-patches and selecting the ones with the highest FOD.
	\item We propose Bounding box Drifting Loss, an IoU-based objective function specialized for patch attacks. In this way, the gradient passed from the regression branch can be stopped if there are no overlaps between the detected Bboxes and the initial ones. 
\end{itemize}
\section{Related Work}
In this section, we first provide a brief review of the classical researches on adversarial attacks for image recognition. Later, we will pay much attention to the studies on adversarial attacks for object detection.
\subsection{Adversarial Attacks on Image Recognition}
Early studies on adversarial vulnerability mainly focus on the task of image recognition. Among them, the white-box setting~\cite{goodfellow2014explaining, szegedy2013intriguing, carlini2017towards, madry2017towards, moosavi2016deepfool}, where the attackers have full access to the victim model, received a wide-spread attention. Later, on the black-box setting, some researchers tried to enhance the transferability via data-augmentation~\cite{ dong2019evading, xie2019improving, lin2019nesterov}, advanced optimization scheme~\cite{dong2018boosting, lin2019nesterov}, etc. Besides, some other attack scenarios also received in-depth studies,~\eg, query-based black-box attacks~\cite{chen2020hopskipjumpattack, brendel2017decision, SUN2022108728, brunner2019guessing} and model stealing~\cite{sun2022exploring, zhou2020dast}. In fact, there is a wide variety of works on this topic. Limited by the space, we could not review them thoroughly in this paper. To this end, we recommend our readers~\cite{silva2020opportunities}~for more comprehensive reviews of the research progress. 
\subsection{Adversarial Attacks on Object Detection}
\begin{figure}[t]
	\centering
	\subfloat[Local loss surface.]{
		\includegraphics[width=.45\linewidth]{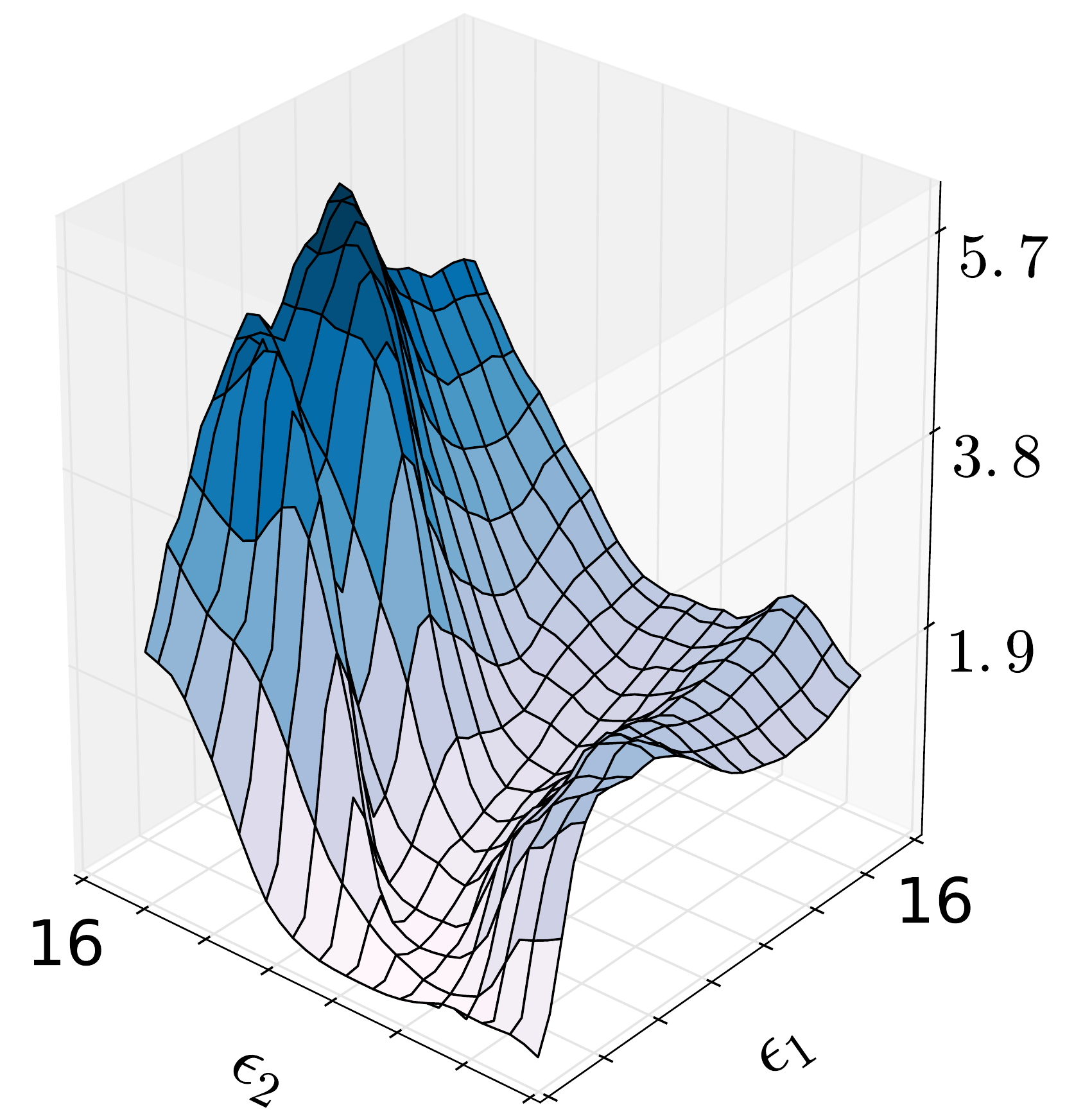}
	}
	\hspace{-4mm}\subfloat[Zoom in of the left figure.]{
		\includegraphics[width=.45\linewidth]{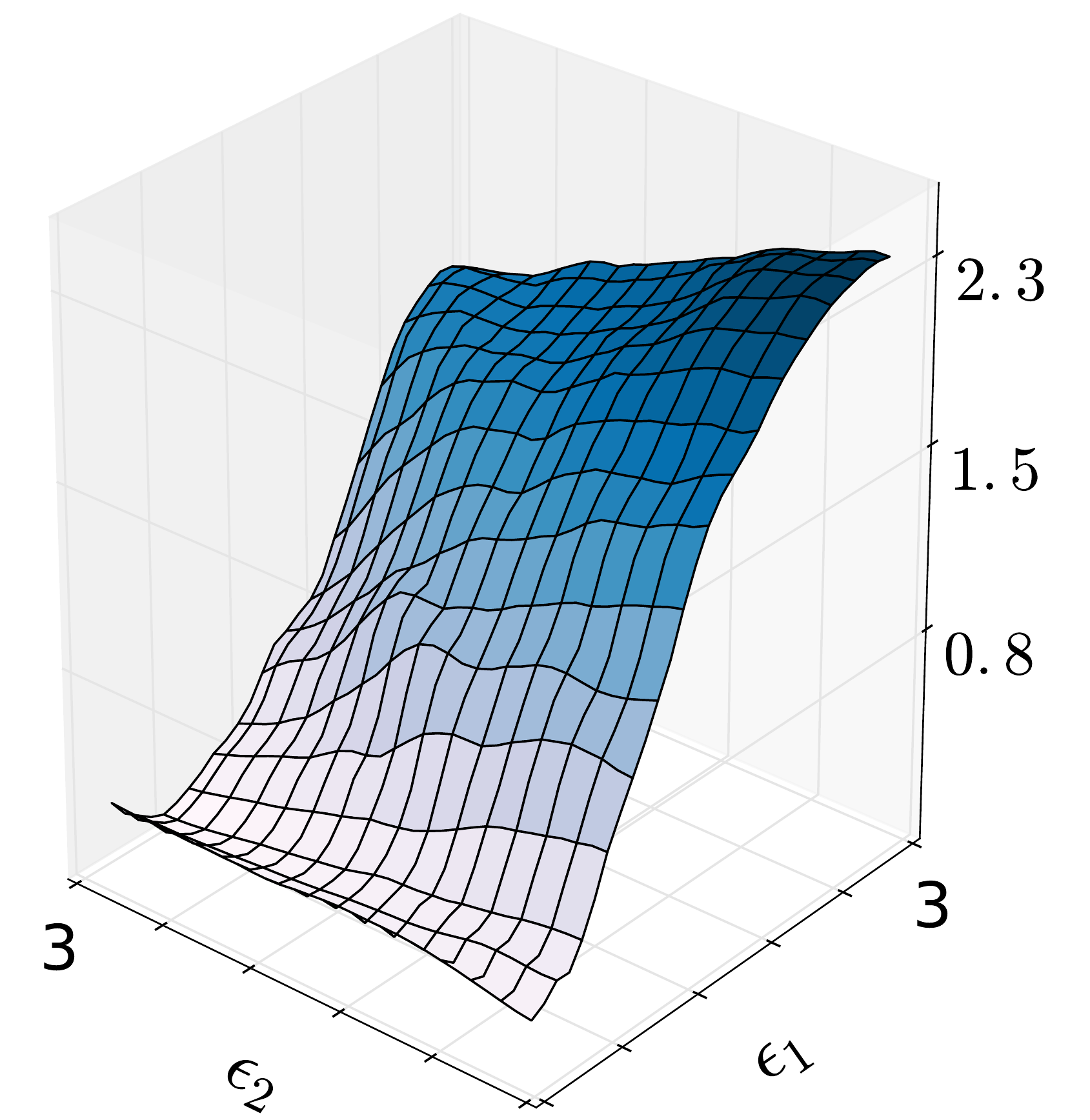}
	}
	\caption{\textbf{Illustrations of the local curvature artifacts~\cite{tramer2018ensemble}.}~It reflects the loss of the adversarially trained Inception-v3~\cite{szegedy2016rethinking}~($ \text{v3}_\text{adv} $), with the form of $ \boldsymbol{x}^*= \boldsymbol{x} + \epsilon_1\cdot\boldsymbol{g}+\epsilon_2\cdot\boldsymbol{g}^\perp$. Here, $ \boldsymbol{g} $ is the signed gradient of $ \text{v3}_\text{adv} $ and $ \boldsymbol{g}^\perp $ is an orthogonal adversarial direction, sampled from Inception-v4~\cite{szegedy2017inception}.}\label{fig2}
\end{figure}
\noindent\textbf{Full-Scale Attacks.}~Xie~\etal~\cite{xie2017adversarial}~propose a first attempt, dubbed DAG, to attack object detection and segmentation. It sets a wrong category for the target to increase the confidence of negative samples while reducing the confidence of positive samples in an iterative manner. Inspired by Carlini \etal~\cite{carlini2017towards} and the Expectation over Transformation~(EoT)~\cite{athalye2018synthesizing}, Chen~\etal~\cite{chen2019shapeshifter}~propose to add random perturbation over each iteration so as to enhance the robustness of adversarial examples for Faster R-CNN~\cite{ren2015faster}. Later, RAP~\cite{lirobust}~attacks the Region Proposal Networks~(RPN)~in two-stage object detectors by destroying both the classification and regression. In addition to the objective function of RAP~\cite{lirobust}, Zhang~\etal~\cite{zhang2020contextual}~introduces contextual loss to increase the confidence of the background and inhibits the confidence of the foreground. More recently, Nezami~\etal~\cite{nezami2021pick}~proposes to precisely manipulate the pixel of the target object to change its label without affecting the other objects. 

\noindent\textbf{Patch Attacks.}~In this field, Wu~\etal~\cite{wu2020dpattack}~proposes a~diffused patch with the shape of asteroid-like or grid-like and pays more attention to the proposals that escaped from attack. Zhao~\etal~\cite{zhao2020object}~designs heatmap-based and consensus-based algorithms to select patches for the attack. Recently, RPAttack~\cite{huang2021rpattack}~ enhances the threats of PAs in patch selection and optimization schemes. Specifically, it proposes a patch selection based on the gradient feedback and leverages the ensemble learning to improve the attack strength.

\noindent\textbf{Adversarial Patches.}~Brown~\etal~\cite{brown2017adversarial}~is the first attempt to attack object detection via a single adversarial patch. It designs an unrestricted patch with a fixed position to attack the classification branch. Liu~\etal~\cite{liu2018dpatch}~mislead the victim detector by forcing it to perceive only the stoked rectangular patch at a fixed position. Liu~\etal~\cite{ liu2019perceptual}~proposed a perceptual-sensitive generative adversarial network to synthesize adversarial patches. Moreover, there are a great variety of researches on adversarial patches, including aerial detection~\cite{lian2022benchmarking, den2020adversarial, lu2021scale}~and physical APs for real-world detection~\cite{thys2019fooling, lee2019physical, wang2021dual}, etc.

In summary, current works regarding adversarial attacks on object detection pay much attention to FSAs and APs, while the research on PAs has not received widespread attention. However, given the threats posed by the imperceptibility of PAs, it deserves in-depth research. Therefore, we take a closer look at PAs on object detection in this paper.

\section{Preliminary}
\subsection{Problem Formulation}
Considering the general representation, we denote the ground truth of an image $\boldsymbol{x}$ as:
\begin{equation}
	\boldsymbol{O}(\boldsymbol{x})=\left\{\boldsymbol{B}_i(\boldsymbol{x}), \boldsymbol{y}_i(\boldsymbol{x})\right\}, \quad(i=1,2,3 \ldots N),
	\label{eq1}
\end{equation}
where $N$ denotes the number of instance in $\boldsymbol{x}$. Here, $\boldsymbol{B}_i(\boldsymbol{x})=\{B^x_i, B^y_i, B^w_i, B^h_i\}$ is the location information with $ (B^x_i, B^y_i)$ denote the coordinates of the Bbox center point, and $ (B^w_i, B^h_i) $ are the width and height of the Bbox. The class information $ \boldsymbol{y}_i(\boldsymbol{x}) \in \{1, 2, ..., C\}$ denotes the label of an instance, where $ C $ is the number of categories. In this way, the detected results for a given image $ \boldsymbol{x} $ can be represented as:
\begin{equation}
	\widetilde{\boldsymbol{O}}(\boldsymbol{x})=\left\{\widetilde{\boldsymbol{B}}_i(\boldsymbol{x}), \widetilde{\boldsymbol{P}}_i(\boldsymbol{x})\right\},
	\label{eq2}
\end{equation}
where $\widetilde{\boldsymbol{P}}_i(\boldsymbol{x})=\{\widetilde{P}^1_i, \widetilde{P}^2_i,...,\widetilde{P}^C_i\}$ is the class probability vector. Then, the detected class is the index corresponding to the maximum in $ \widetilde{\boldsymbol{P}}_i(\boldsymbol{x}) $. That is,
\begin{equation}
	\widetilde{C}_i(\boldsymbol{x}) = \text{argmax}\widetilde{\boldsymbol{P}}_i(\boldsymbol{x}).
	\label{eq3}
\end{equation}

Based on the above notations, the objective of adversarial attacks on object detection can be formulated as the following:
\begin{equation}
\begin{aligned}
		&\min\ ||\boldsymbol{\xi}||_p\\
		 s.t.\ \boldsymbol{O}&\left(\boldsymbol{x}_{\text{adv}}\right) \neq \boldsymbol{O}(\boldsymbol{x})
\end{aligned}\quad,
\label{eq4}
\end{equation}
where $ \boldsymbol{x} $ are clean images and $ \boldsymbol{\xi} $ are the corresponding perturbations. The objective in~\cref{eq4} is to find the adversarial example with the minimum visual distortion. Since both our method and our competitors are based on the BIM optimizing framework, we choose the $ \ell_\infty $ norm as the visual constraint,~\ie, $ p=\infty $ in~\cref{eq4}.

Here, the misleading to either the branch of classification or regression can be seen as a successful attack. That is, either $ \widetilde{C}_i(\boldsymbol{x}) \neq \widetilde{C}_i(\boldsymbol{x}_\text{adv})  $ or $ \text{IoU}(\widetilde{\boldsymbol{B}}_i(\boldsymbol{x}), \widetilde{\boldsymbol{B}}_i(\boldsymbol{x}_\text{adv}))<0.5  $ can be seen as a successful attack. Besides, for PAs, the representation of adversarial examples can be formulated as:
\begin{equation}
	\boldsymbol{x}_\text{adv}  = \boldsymbol{x} +  \textbf{M}^*\odot\boldsymbol{\xi}^*,
	\label{eq5}
\end{equation}
where $ \odot $ is the element-wise multiplication. $ \textbf{M} $ is the attack map with the same size as $ \boldsymbol{x} $, which determine the regions to be attacked. Here, $ \boldsymbol{\xi}^* $ represents the optimal solution of~\cref{eq4} and $ \textbf{M}^* $ denotes the optimal results of the attack map, which is fixed over the entire attack progress in general.
\begin{figure}[t]\setcounter{figure}{2}
	\centering
	\includegraphics[scale=1.0]{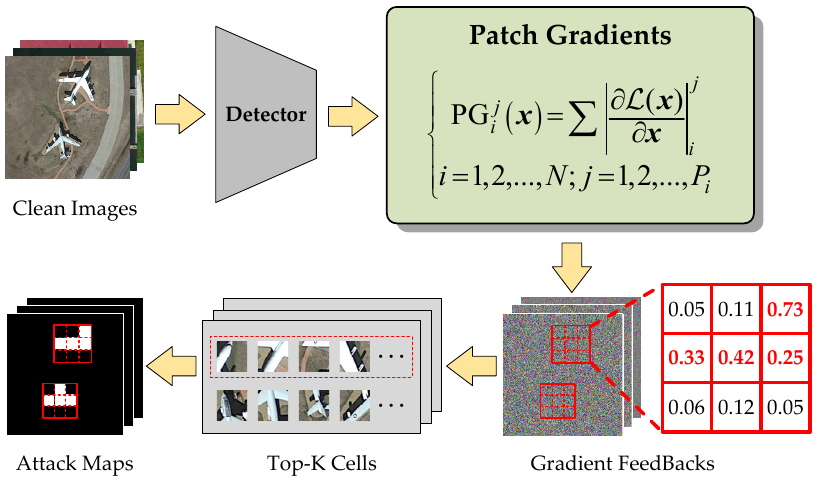}
	\caption{\textbf{Illustration of the patch selection scheme in RPAttack~\cite{huang2021rpattack}.}}
	\label{fig3}
\end{figure}
\subsection{Patch Selection Scheme}
\begin{figure*}[thbp]\setcounter{figure}{3}
	\centering
	\includegraphics[scale=1.]{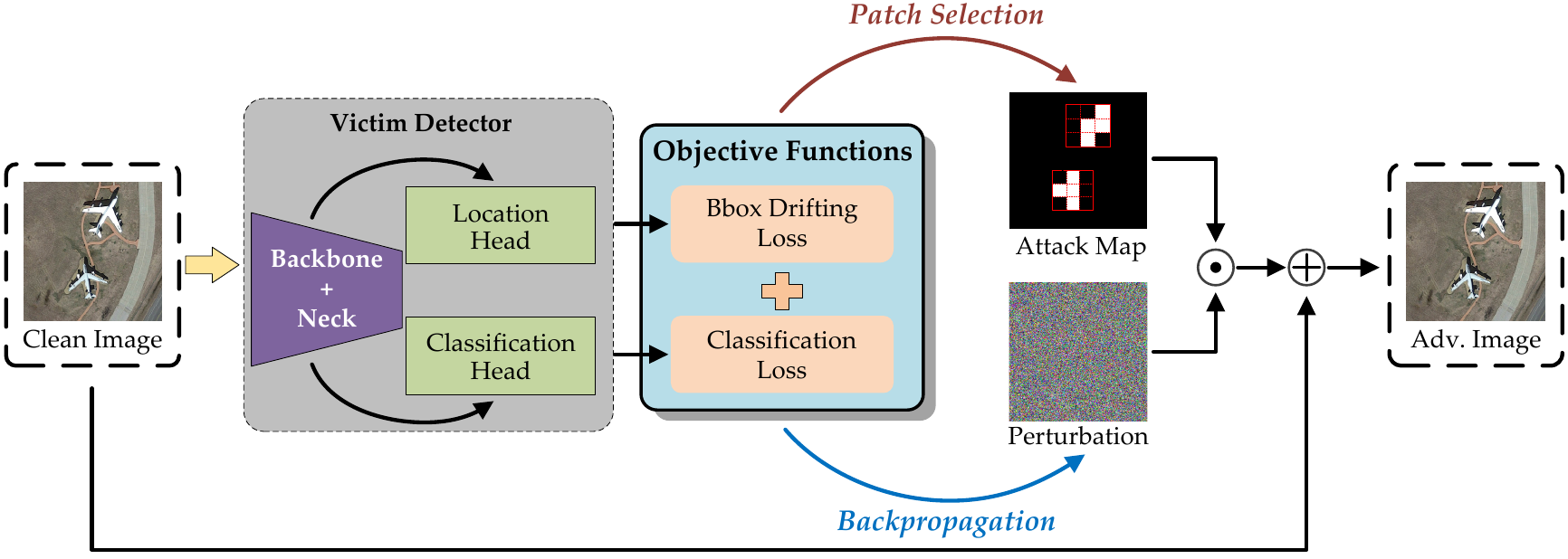}
	\caption{\textbf{The framework of our proposed TPA.}~Here, $ \odot $ and $ \oplus $ denote the element-wise multiplication and addition, respectively. Also, the generated adversarial examples should be clipped to legal interval of image, as shown in~\cref{eq5}. Please refer to~\cref{sec4-1} for more exact illustration.}
	\label{fig4}
\end{figure*}
In this section, we mainly introduce the patch selection scheme in RPAttack~\cite{huang2021rpattack}. For simplicity, we dub it Gradient Feedback~(GF). As can be seen in~\cref{fig3}, it first splits the Bbox of each instance evenly to get a series of sub-patches. Then, the sub-patches are ranked~\wrt~the gradient norm, and the top-k sub-patches with  the largest gradient norms are selected. Here, $ \mathcal{L} $ denotes the objective function leveraged for the optimization of the perturbations and $ \text{PG}^j_i(\boldsymbol{x}) $ denotes the $ \ell_1 $ norm of the $ j $-th sub-patch in the $ i $-th instance of an image $ \boldsymbol{x} $.
\subsection{Objective Function}
Here, the objective function proposed by~\cite{lirobust}~to attack the regression branch of a detector is introduced in this section. For simplicity, we dub it the Coordinate-Based Loss~(CBL). In general, attacking the regression branch aims at making the detected Bboxes own undesirable shape or position. To this end, RAP assigns large offsets for the detected Bboxes to make them cover the entire image as much as possible. Formally, the CBL is expressed as:
\begin{equation}
	\begin{aligned}
		\mathcal{L}_\text{CBL}(\widetilde{\boldsymbol{B}}(\boldsymbol{x}_\text{adv})) &= \sum^m_{j=1} z_j((B_j^x-\tau^x)^2 + (B_j^y-\tau^y)^2\\
		& + (B_j^w-\tau^w)^2 + (B_j^h-\tau^h)^2)
	\end{aligned}\quad,
	\label{eq6}
\end{equation}
where $ \{\tau^x, \tau^y, \tau^w, \tau^h\} $ is the predefined offsets. Besides, $ z_j $ is the indicator of $ j $-th proposal and its formulation can be expressed as:
\begin{equation}
\left\{
\begin{aligned}
&z_j = 1, \text{if}\ 
\left\{
\begin{aligned}
	&\text{IoU}(\widetilde{\boldsymbol{B}}_j(\boldsymbol{x}), \widetilde{\boldsymbol{B}}_j(\boldsymbol{x}_\text{adv})) > 0.1\\
	&\text{max}\widetilde{\boldsymbol{P}}_j(\boldsymbol{x}_\text{adv}) > 0.4
\end{aligned}\right.\\
&z_j = 0, \text{otherwise}
\end{aligned}\right.\ .
\label{eq7}
\end{equation}

In this paper, we follow the setting of RAP~\cite{lirobust} to set $ \tau^x = \tau^y = \tau^w = \tau^h = 10^5$ in~\cref{eq7}.
\section{Method}
\subsection{Overview}\label{sec4-1}
This paper aims to propose a threatening patch attack that is applicable to multiple target detectors, including Faster R-CNN~\cite{ren2015faster}, RetinaNet~\cite{lin2017focal}, YOLO-v4~\cite{bochkovskiy2020yolov4}, and FCOS~\cite{tian2019fcos}. Since the structures of these detector networks are different, the attack framework is not designed specifically for a certain kind of detector but based on the final prediction results. Here, we depict the framework of our TPA in~\cref{fig4}. Specifically, we first select the regions to acquire the attack map $ \textbf{M} $ via the proposed FOD patch selection scheme, and $ \textbf{M} $ will be fixed over the entire attack progress. Then, similar to BIM~\cite{kurakin2018adversarial}, we optimize the adversarial example in an iterative manner.

Formally, at each iteration $ t+1 $, we have:
\begin{equation}
\begin{aligned}
\boldsymbol{x}_\text{adv} &= \hat{\boldsymbol{x}}_{t+1}\\
& = \boldsymbol{x} + \textbf{M}\odot\boldsymbol{\xi}_{t+1}
\end{aligned}\quad,
\label{eq8}
\end{equation}
where $ \boldsymbol{\xi}_{t+1} $ is formulated as:
\begin{equation}
	\boldsymbol{\xi}_{t+1} = \text{Clip}^\epsilon_{-\epsilon}\{\text{Clip}^1_0\{\hat{\boldsymbol{x}}_{t} + \textbf{M}\odot [\alpha \cdot \text{sign}(\boldsymbol{g}_{t+1})]\}-\boldsymbol{x}\},
	\label{eq9}
\end{equation}
where $ \epsilon $ denotes the $ \ell_\infty $ constraint and $ \alpha $ is the attack step size. In our TPA, the attack map $ \textbf{M} $ is a binary mask, which determines where to attack. Besides, the expression of $ \boldsymbol{g}_{t+1} $ in~\cref{eq9} is:
\begin{equation}
\boldsymbol{g}_{t+1} = \frac{\nabla_{\hat{\boldsymbol{x}}_{t}}\mathcal{L}(\hat{\boldsymbol{x}}_{t})}{||\nabla_{\hat{\boldsymbol{x}}_{t}}\mathcal{L}(\hat{\boldsymbol{x}}_{t})||_1}.
\label{eq10}
\end{equation}
 Here, the specific details regarding how to get the attack map $ \textbf{M} $ and the expression of $ \mathcal{L}(\boldsymbol{x}) $ will be introduced next.
\subsection{First-Order Difference Patch Selection Scheme}\label{sec4-2}
For patch attacks, the selection of sub-patches is a critical factor that affects the attack efficiency. However, as we have mentioned in~\cref{sec1}, the advanced patch selection scheme proposed by RPAttack~\cite{huang2021rpattack} may suffer from the problem of inconsistency between the local and global landscapes, since the gradient is defined within a small neighborhood around the data point, which is not large enough to explore the global landscapes. To this end, we propose to imitate the ``masking" manipulation in patch attacks by covering each sub-patch of the instance and select the sub-patches with the highest feedback.

Specifically, as shown in~\cref{fig5}, we first divide the Bbox of each instance into a grid of $ n\times n $. Here, considering the objects in O-RSIs with the same class could have different sizes, $ n $ can vary according to the size of the instance. In this paper, we provide two options for the grid segmentation,~\ie,~the uniform segmentation scheme and the scale-adaptive segmentation scheme. All the instances in the uniform segmentation share the same setting of $ n $. For simplicity, we denote $ \text{U}(n) $ as the uniform segmentation scheme with the size of $ n $. For the scale-adaptive segmentation scheme, similar to MS COCO~\cite{lin2014microsoft}, we divide the instances into three kinds of scales in terms of their areas. To be specific, we denote the area of an instance as $ \mathcal{S} $. Then, we set $ n $ to $ n_1 $ for $ \mathcal{S}\leq 32^2 $, $ n_2 $ for $ 32^2<\mathcal{S}\leq64^2 $, and $ n_3 $ for $ \mathcal{S}>64^2 $. Thus, we can use $ \text{SA}(n_1, n_2, n_3) $ to represents the scale-adaptive segmentation scheme with certain parameters. The results regarding these two patch segmentation scheme will be reported and analyzed in~\cref{sec5-3}.

Once we acquire the masked inputs, we feed them into the victim detector to calculate the FOD. Formally, for a sub-patch, we define the FOD as:
\begin{equation}
\begin{aligned}
	\text{FOD}^j_i(\boldsymbol{x})&= \mathcal{L}(\boldsymbol{x}) - \mathcal{L}(\boldsymbol{x}^j_i)\\
	&=\mathcal{L}(\boldsymbol{x}) - \mathcal{L}(\textbf{M}^j_i\odot\boldsymbol{x})
\end{aligned}\quad,
\label{eq11}
\end{equation}
where $ \textbf{M}^j_i $ denotes masking the $ j $-th sub-patch of the $ i $-th instance. Correspondingly, $ \boldsymbol{x}^j_i $ is the input with the $ j $-th sub-patch of the $ i $-th instance is masked, and $ \text{FOD}^j_i(\boldsymbol{x}) $ is the first-order difference of the $ j $-th sub-patch belonging to the $ i $-th instance. Associating with the objective function utilized in our TPA~(please refer to~\cref{sec4-3} for more detailed introduction),~\cref{eq11} can be expressed as:
\begin{equation}
	(\text{max}\widetilde{\boldsymbol{P}}_i(\boldsymbol{x}) - \widetilde{\boldsymbol{P}}^{\widetilde{C}_i(\boldsymbol{x})}_i(\boldsymbol{x^j_i})) 
	+ (1 - \text{IoU}(\widetilde{\boldsymbol{B}}_i(\boldsymbol{x}), \widetilde{\boldsymbol{B}}_i(\boldsymbol{x}^j_i))),
	\label{eq12}
\end{equation}
where $ \widetilde{\boldsymbol{P}}^{\widetilde{C}_i(\boldsymbol{x})}_i(\boldsymbol{x^j_i}) $ denotes the $ \widetilde{C}_i(\boldsymbol{x}) $-th entry of $ \widetilde{\boldsymbol{P}}_i(\boldsymbol{x^j_i}) $.
 \begin{figure*}[thbp]\setcounter{figure}{4}
	\centering
	\includegraphics[scale=1.]{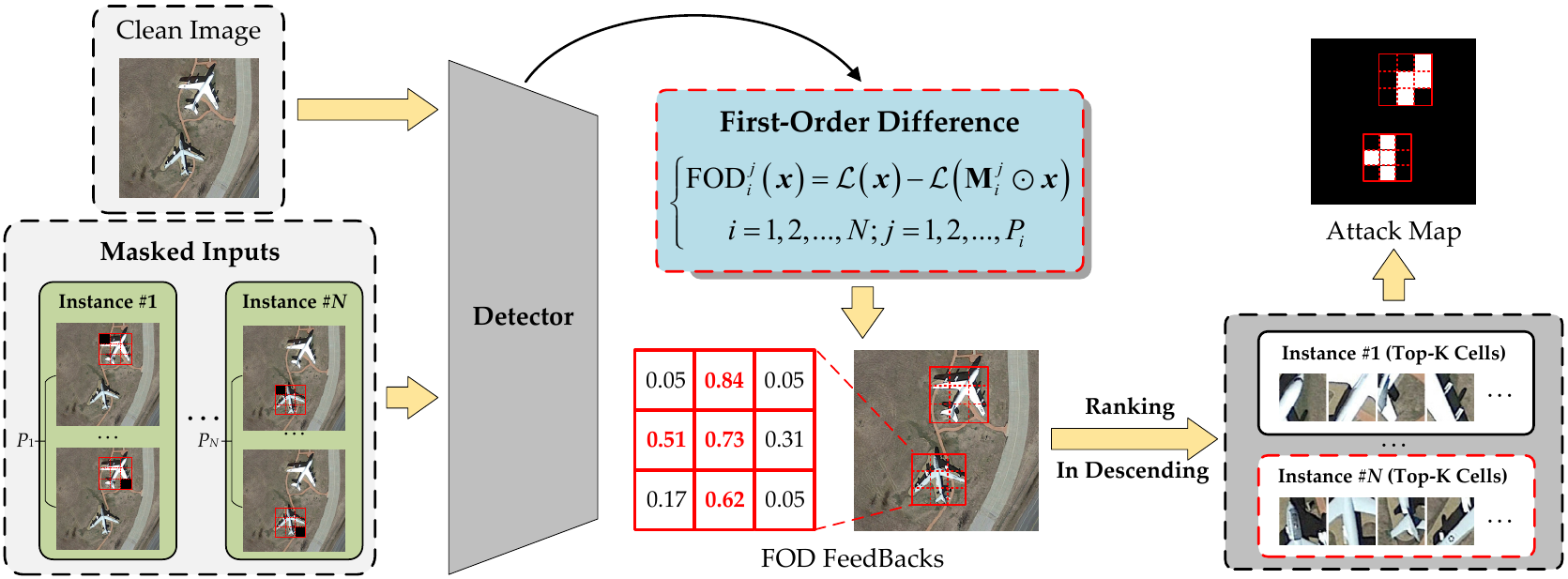}
	\caption{\textbf{Illustration of the proposed FOD patch selection scheme.}~}
	\label{fig5}
\end{figure*}
\subsection{Bounding box Drifting Loss}\label{sec4-3}
Compared to the image classifier, attacking the object detector is deemed as a more complicated problem, on account of the complex outputs from the detector. Therefore, in addition to attacking the classification branch, destroying the regression branch of the detector is another essential step, which could enhance the threats of attacks to a large extent. Existing methods for PAs only focus on the attack against the classification branch. Fortunately, RAP~\cite{lirobust}~has proposed CBL to break the Bbox regression branch. However, as we analyzed in~\cref{sec1}, applying CBL directly to the framework of PAs may be at risk of gradient inundation. In other words, since PAs only attack some specific regions of an image, making all the detected Bboxes covering the entire image seems sets an \textit{impossible} objective. As a result, the gradients of CBL will exist over the entire attack progress. Besides, the norms of the gradient passed from CBL are larger than that passed from the classification loss, on account of the large threshold in CBL. Consequently, the gradient of the classification loss may suffer from being inundated by that of CBL, leading to the stagnation of optimization.

In fact, with the goal of attacking the regression branch, offsetting the detected Bboxes is exact what we want to see. To this end, there exist many solutions for this purpose. One of the most threatening scenario is that there are no intersections between the initial Bbox and detected ones after attacking. Thus, we formulate the attacking on the regression branch as the above situation. That is, drifting the detected Bboxes away from the initial one until there are no overlap between them. When it comes to measuring the overlaps between two Bboxes, a natural idea is to leverage the IoU, a commonly-adopted metric in object detection. Therefore, we formulate our Bbox drifting loss as:
\begin{equation}
\mathcal{L}_\text{BDL}(\hat{\boldsymbol{x}}) = \frac{1}{N}\sum^N_{i=1}\text{max}(\text{IoU}(\widetilde{\boldsymbol{B}}_i(\boldsymbol{x}), \widetilde{\boldsymbol{B}}_i(\hat{\boldsymbol{x}}))) ,
\label{eq13}
\end{equation}
where $ N $ denotes the number of instances in the initial results. During the iterations, there may exist a lot of detected Bboxes around the initial one. To this end,~\cref{eq13} takes the Bbox with the highest IoU between the initial one into calculation. In this way, the detected Bboxes that have no overlaps between the initial one are not taken into consideration. That is, these Bboxes have been attacked successfully.

Another loss function utilized in our TPA is for the attack on the classification branch. Here, we use the loss in RPAttack~\cite{huang2021rpattack}, which is formulated as:
\begin{equation}
\mathcal{L}_\text{cls}(\hat{\boldsymbol{x}}) = \frac{1}{k}\sum^k_{i=1}||\text{max}\widetilde{\boldsymbol{P}}_i(\hat{\boldsymbol{x}})||^2,
\label{eq14}
\end{equation}
where $ k $ denotes the number of detected results in $ \hat{\boldsymbol{x}} $. Finally, we use~\cref{eq15} as the total objective function in our TPA.
\begin{equation}
	\mathcal{L}(\hat{\boldsymbol{x}}) = \mathcal{L}_\text{BDL}(\hat{\boldsymbol{x}}) + \mathcal{L}_\text{cls}(\hat{\boldsymbol{x}}).
	\label{eq15}
\end{equation}
\section{Experiments}
\subsection{Experimental Settings}
\begin{figure}
	\centering
	\subfloat[DIOR v.s. DIOR-A w.r.t. Class-Wise Instance Distribution.]{
		\includegraphics[width=.47\textwidth]{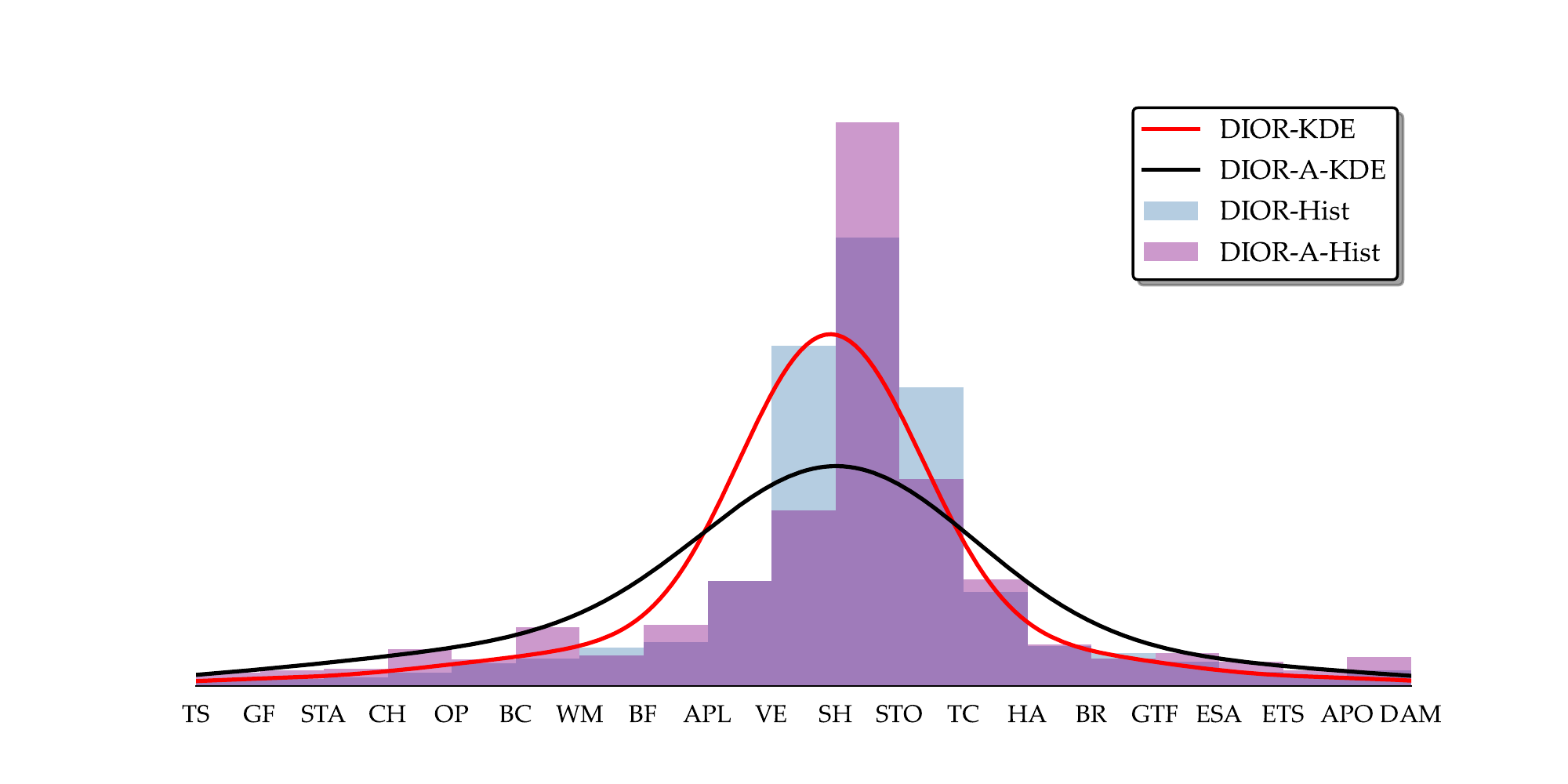}
	}\\
	\vspace{-1mm}
	\subfloat[DOTA v.s. DOTA-A w.r.t. Class-Wise Instance Distribution.]{
		\includegraphics[width=.47\textwidth]{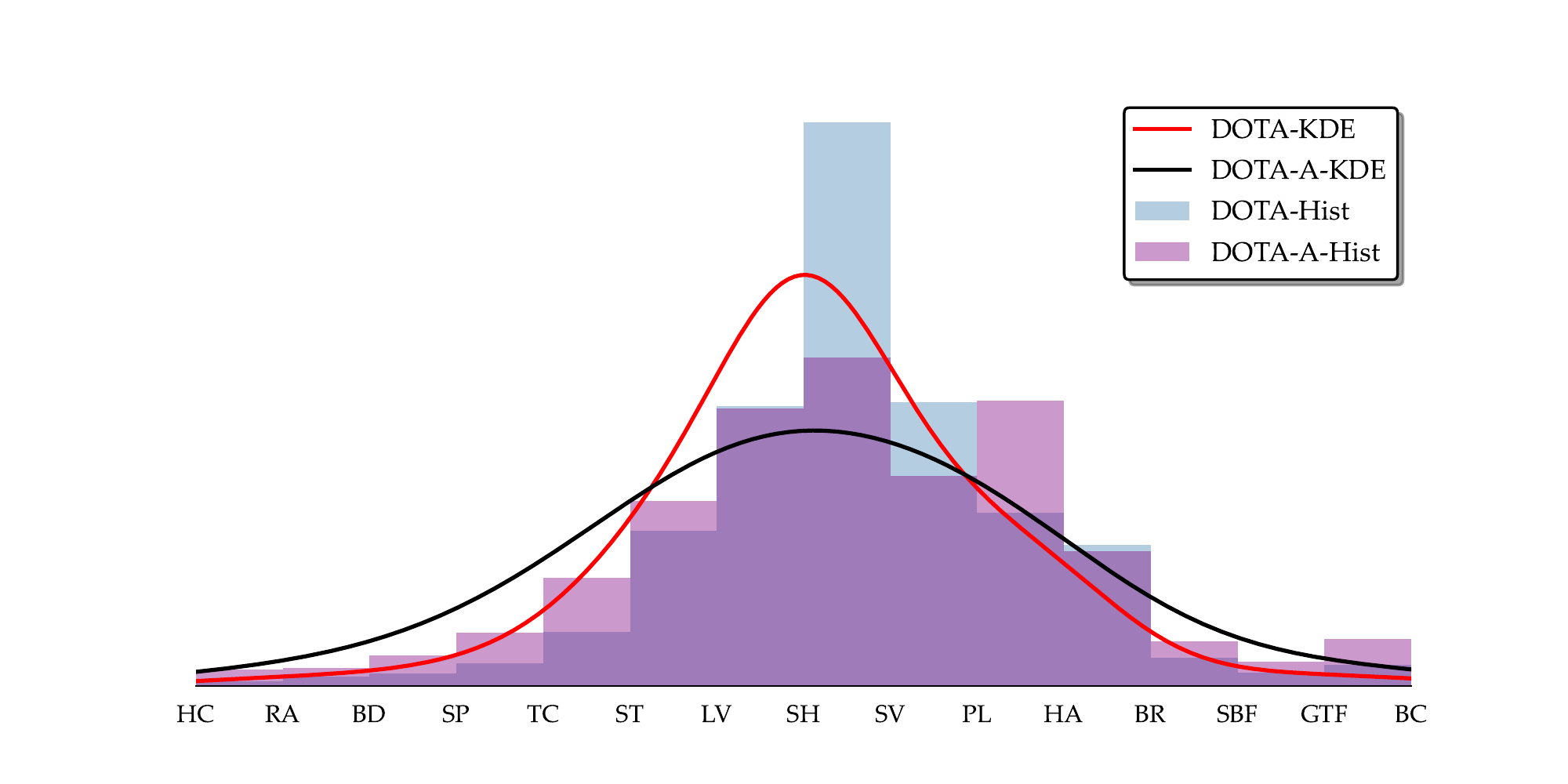}
	}
	\caption{\textbf{Class-Wise instance distributions.}}
	\label{fig6}
\end{figure}

\noindent\textbf{Datasets.}~We carry out the evaluations on two widely-adopted benchmarks for object detection in O-RSIs,~\ie,~DIOR~\cite{cheng2021anchor} and DOTA~\cite{xia2018dota}. For DIOR dataset, it contains 23463 images in RGB color space, covering 192518 instances of 20 categories. All the images are formatted in a fixed size of $ 800\times 800 $ with a spatial resolution varying across 0.5 to 30 meters. DOTA dataset (we use DOTA-1.0 in this paper) includes 2860 images covering 15 categories, and the size of images in DOTA vary across $ 800\times800 $ to $ 4000\times4000 $. In practical, considering computational burden caused by the large scale of images in DOTA, we split the images into a fixed size, which is set to $ 800\times800 $ in this paper. Besides, to facilitate the evaluation of the following research regarding adversarial attacks on object detection in O-RSIs, similar to the commonly-adopted protocol~\cite{dong2018boosting, huang2021rpattack}~in the field of adversarial attacks, we sample 2000 images from the testing subset of DIOR and the validation subset of DOTA, respectively, dubbed DIOR-A and DOTA-A. The class-wise instance distributions of them are exhibited in~\cref{fig6}. Here, we plot both the class-wise instance distribution histograms and their corresponding Kernel Density Estimation (KDE) curves, in which we can see that the sampled datasets share almost the same class-wise distribution with their corresponding parent datasets. Besides, since only 2000 images are utilized for the evaluation, they could reduce the calculation budget to some extent, compared to using the parent datasets for the evaluation. Thus, we will carry out all the following experiments on these sampled datasets.
\begin{table*}[tbp]\setcounter{table}{1}
	\centering
	\caption{\textbf{Comparison Results on DIOR-A and DOTA-A Datasets.}~Results are separated by the double line, above which are the results on DIOR-A and the remains are on DOTA-A. The best results are shown in \textbf{bold}.}
	\setlength{\tabcolsep}{1.5mm}{
		\begin{threeparttable}
			\begin{tabular}{c|cc|cc|cc|cc|cc|cc|cc}
				\toprule
				Detector & \multicolumn{2}{c|}{FR-50} & \multicolumn{2}{c|}{FR-101} & \multicolumn{2}{c|}{FC-50} & \multicolumn{2}{c|}{FC-101} & \multicolumn{2}{c|}{RT-50} & \multicolumn{2}{c|}{RT-101} & \multicolumn{2}{c}{YOLO-v4} \\
				\midrule
				Method & RPA~\cite{huang2021rpattack} & Ours  & RPA~\cite{huang2021rpattack} & Ours  & RPA~\cite{huang2021rpattack} & Ours  & RPA~\cite{huang2021rpattack} & Ours  & RPA~\cite{huang2021rpattack} & Ours  & RPA~\cite{huang2021rpattack} & Ours  & RPA~\cite{huang2021rpattack} & Ours \\
				\midrule
				mAP ($\%$) & 53.20  & \textbf{47.80} &   55.80    & \textbf{50.80}  &  -  & \textbf{47.00} & -  & \textbf{51.90} & - & \textbf{39.60} & - & \textbf{44.70} & 50.40 & \textbf{20.30} \\
				Recall ($\%$) & 69.70  & \textbf{55.80} & 71.00  & \textbf{58.30}  & - & \textbf{61.50} & - & \textbf{66.40} & - & \textbf{58.00} & -  & \textbf{60.70} & 29.60 & \textbf{20.20} \\
				$\ell_0$ Norm & 0.110  & \textbf{0.059} &  0.120  & \textbf{0.059}  & -  & \textbf{0.059} & -  & \textbf{0.059} & -  & \textbf{0.059} & -  & \textbf{0.059} & 0.110  & \textbf{0.059} \\
				$\ell_2$ Norm & 6.840  & \textbf{4.990} & 6.860  & \textbf{5.160}  & -  & \textbf{5.640} & -  & \textbf{5.790} & -  & \textbf{5.170} & -  & \textbf{5.310} & 6.840  & \textbf{4.860} \\
				\midrule
				\midrule
				mAP ($\%$) & 31.90  & \textbf{26.60} &   54.00    & \textbf{39.40}  &  -  & \textbf{28.70} & -  & \textbf{32.10} & - & \textbf{17.10} & - & \textbf{20.50} & 34.80 & \textbf{21.70} \\
				Recall ($\%$) & 49.80  & \textbf{26.60} & 65.50  & \textbf{48.20}  & - & \textbf{47.60} & - & \textbf{50.30} & - & \textbf{37.00} & -  & \textbf{40.50} & 36.40 & \textbf{22.30} \\
				$\ell_0$ Norm & 0.060  & \textbf{0.032} &  0.060  & \textbf{0.031}  & -  & \textbf{0.033} & - & \textbf{0.032} & - & \textbf{0.032} & - & \textbf{0.032} & 0.060  & \textbf{0.032} \\
				$\ell_2$ Norm & 5.412  & \textbf{3.628} & 5.557  & \textbf{3.756}  & - & \textbf{4.304} & - & \textbf{4.411} & - & \textbf{3.944} & - & \textbf{4.005} & 5.412  & \textbf{3.731} \\
				\bottomrule
			\end{tabular}
			\label{tab2}%
			\footnotesize Since RPAttack~\cite{huang2021rpattack}~(RPA) utilizes Faster R-CNN~\cite{ren2015faster} and YOLO-v4~\cite{bochkovskiy2020yolov4} to carry out an ensemble attack, it can not attack RetinaNet and FCOS under the white-box setting. In this case, we only report the results of RPAttack on FC-50, FC-101, and YOLO-v4.
	\end{threeparttable}}
\end{table*}%

\noindent\textbf{General settings.}~We leverage Pytorch framework to implement our method on a single NVIDIA RTX 2080Ti. Here, four kinds of detectors are utilized for the evaluations, where Faster R-CNN~\cite{ren2015faster}~(FR), RetinaNet~\cite{lin2017focal}~(RT), and FCOS~\cite{tian2019fcos}~(FC) are trained on MMDetection, and YOLO-v4~\cite{bochkovskiy2020yolov4} are based on DarkNet~\cite{redmon2016you}. For the detectors on MMDetection, we equip them with two backbones including ResNet-50~\cite{he2016deep} with FPN~\cite{lin2017feature}~and ResNet-101~\cite{he2016deep} with FPN~\cite{lin2017feature}. For simplicity, FR-50 denotes Faster R-CNN with ResNet-50+FPN as the backbone, and so on. The testing results on clean images of DIOR-A and DOTA-A are reported in~\cref{tab1}. Here, for DIOR-A, we utilize the train-val subsets to train the victim detectors, and training subset is leveraged to train the detectors for DOTA-A. More training details and the sampled datasets are open accessed\footnote{\href{https://github.com/plpl2019/TPA}{https://github.com/plpl2019/TPA}}.
\begin{table}[tbp]\setcounter{table}{0}
	\centering
	\caption{\textbf{Results on DIOR-A and DOTA-A datasets.}}
	\setlength{\tabcolsep}{0.9mm}{
		\begin{threeparttable}
			\begin{tabular}{c|ccccccc}
				\toprule
				Detector & FR-50 & FR-101 & FC-50 & FC-101 & RT-50 & RT-101 & YOLO-v4\\
				\midrule
				mAP ($\%$) & 88.30 & 88.60 & 87.30 & 87.60 & 87.30 & 87.30 & 89.50\\
				Recall ($\%$) & 90.30 & 90.90 & 91.30 & 91.60 & 92.80 & 92.80 & 90.00\\
				\midrule
				\midrule
				mAP ($\%$) & 68.70 & 68.40 & 65.70 & 66.80 & 62.20 & 64.80 & 69.70\\
				Recall ($\%$) & 77.70 & 76.10 & 79.10 & 80.00 & 79.50 & 81.30 & 76.80\\
				\bottomrule
			\end{tabular}%
			\label{tab1}%
			\footnotesize Here, the results above the double lines are those on DIOR-A dataset and the others are those on DOTA-A dataset, the same to~\cref{tab2}
	\end{threeparttable}}
\end{table}%

\noindent\textbf{Attack settings.}~Considering the implementation and relativity, we choose RPAttack~\cite{huang2021rpattack}, the state-of-the-art patch attack on object detection in natural images, as our competitor. For the evaluation metrics, we leverage mAP and Recall to measure the strength of different attacks. Besides, we introduce $ \ell_2 $ and $ \ell_0 $ norms to evaluate the visual quality of adversarial examples. For the attack settings, the $ \ell_\infty $ constraint $ \epsilon $ in~\cref{eq9} is set to $ 10/255 $ and the number of iterations $ T $ is $ 10 $. The step size $ \alpha $ in~\cref{eq9} is set to $ 1/255 $. Furthermore, both RPAttack and our TPA attack $ \lfloor \frac{n\times n}{2} \rfloor $ patches of an instance, for which we divide to $ n\times n $ patches in total. Here, $ \lfloor \cdot \rfloor $ represents the floor division. Finally, for the grid segmentation, we use $\text{SA}(1, 2, 3)$ as the main scheme. The reason for this choice will be discussed in~\cref{sec5-3}.

\subsection{Peer Comparisons}
In this subsection, we carry out experiments to validate the advancement of our TPA. The quantitative results are summarized in~\cref{tab2}. Surprisingly, the performance of our TPA can surpass the advanced competitor, RPAttack, which leverages the ensemble setting to enhance its threats, where ensemble setting is a powerful attack setting that utilizes more than one victim for the optimization of adversarial examples. Specifically, we summarize the advantages of our TPA as three folds. First, we exceed RPAttack by a large margin in terms of Recall, which indicates that TPA owns the great potential of hiding targets than RPAttack. Second, we can keep the highest attack strength while keeping the lowest $ \ell_2 $ norms, which demonstrates the great attack efficiency of our TPA. Finally, different from RPAttack that is restricted by the choice of the victim detector, the proposed TPA is applicable to a variety of detection models.

\begin{figure*}[th]
	\centering
	\includegraphics[width=\textwidth]{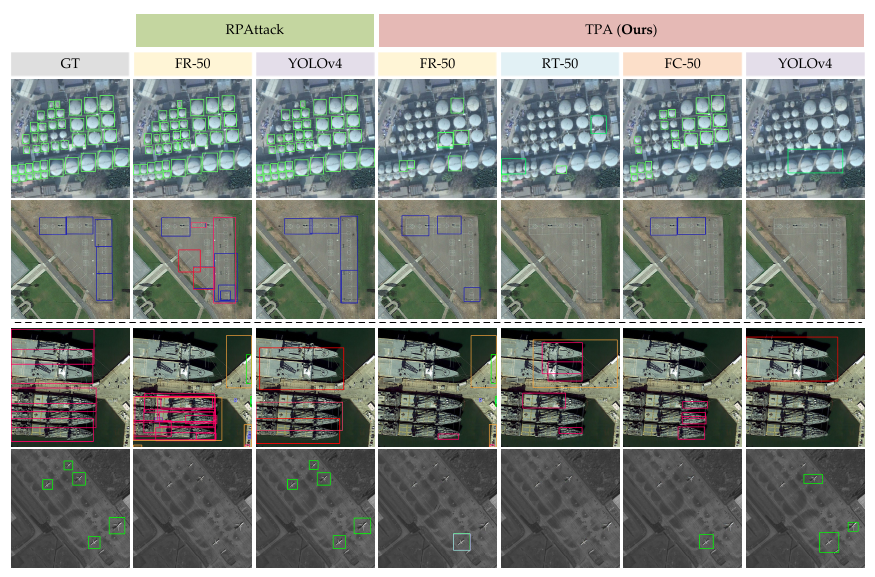}
	\caption{\textbf{Qualitative results w.r.t. the effect of the bounding box drifting loss.}~Here, GT for ground truth. Samples extracted from DIOR-A and DOTA-A are separated by the dotted line, where the upper samples belong to DIOR-A and the remains belong to DOTA-A.}
	\label{fig7}
\end{figure*}

Later, we also visualize the detected results in~\cref{fig7}. Here, all the targets surrounded by the boxes with the same color belong to the same category, and we use different colors to distinct the categories. As we can see from~\cref{fig7}, compared to RPAttack, TPA can make most targets "invisible". In addition, \cref{fig7} releases a significant trend that the invisible ability of RPAttack gets decreased as the density of the targets increases. From this point of view, TPA shows more threats than RPAttack, since the dense instance distribution is a very common phenomenon in O-RSIs.
\begin{table}[tbp]\setcounter{table}{2}
	\renewcommand\arraystretch{1.5}
	\centering
	\caption{\textbf{Ablation Results Regarding The Patch Selection Scheme.}}
	\footnotesize
	\setlength{\tabcolsep}{0.8mm}{
		\begin{tabular}{>{\columncolor{white}}cc|cccc}
			\toprule
			Detector & Method & mAP ($\%$) & Recall ($\%$) &  $\ell_0$ Norm &  $\ell_2$ Norm \\
			\midrule
			& RD & 52.80  & 61.30  & \textbf{0.059}  & 5.050  \\
			& GF~\cite{huang2021rpattack} & 50.00  & 59.40  & \textbf{0.059}  & 4.990  \\
			\rowcolor{gray!20}\cellcolor{white}\multirow{-3}{*}[-0.25ex]{FR-50} & FOD & \textbf{47.80} & \textbf{55.80} & \textbf{0.059} & \textbf{4.900} \\
			\midrule
			& RD & 52.30  & 67.50  & \textbf{0.059}  & 5.640  \\
			& GF~\cite{huang2021rpattack} & 49.60  & 65.00  & \textbf{0.059}  & 5.620  \\
			\rowcolor{gray!20}\cellcolor{white}\multirow{-3}{*}[-0.25ex]{FC-50} & FOD & \textbf{47.00} & \textbf{61.50} & \textbf{0.059} & \textbf{5.510} \\
			\midrule
			& RD & 45.20  & 64.90  & \textbf{0.059}  & 5.180  \\
			& GF~\cite{huang2021rpattack} & 43.10  & 62.90  & \textbf{0.059}  & 5.180  \\
			\rowcolor{gray!20}\cellcolor{white}\multirow{-3}{*}[-0.25ex]{RT-50} & FOD & \textbf{39.60} & \textbf{58.00} & \textbf{0.059} & \textbf{5.170} \\
			\bottomrule
	\end{tabular}}%
	\label{tab3}%
\end{table}%
\subsection{Further Studies}\label{sec5-3}
In this subsection, we arrange extended experiments to take a closer look at our TPA. Specifically, we provide three ablation studies regarding the choice of patch selection scheme, the choice of the objective functions for the Bbox regression, and the choice of patch segmentation scheme.

\noindent \textbf{Ablation Study on Patch Selection Scheme.}~As we have introduced in~\cref{sec1}, RPAttack utilizes the gradient feedback to select the most critical regions to be attacked, which may suffer from the inconsistency between local and global landscapes, leaving the attack efficiency to be suppressed. To this end, we vary the choice of the patch selection scheme in our TPA to see the effect of our FOD. The results of this part are summarized in~\cref{tab3}. Here, GF stands for the gradient feedback patch selection scheme that is leveraged in RPAttack~\cite{huang2021rpattack} and RD represents selecting the regions in a random manner. Not surprisingly, FOD achieves the best results in terms of both strength and visual quality, and picking the sub-patches in a random manner exhibits the poorest performance. That is, selecting the sub-patches indeed plays a critical role in final results. Besides, as we have pointed out in~\cref{sec1}, FOD selects the most critical sub-patches via imitating the attack scheme in PAs so that to escape from the sub-optimal within the local neighborhood to find the sub-patches with the most attack potential. From this point of view, a relatively constrictive neighborhood could mislead the choice of sub-patches, resulting in poor threats.
\begin{table}[tbp]
	\renewcommand\arraystretch{1.4}
	\centering
	\caption{\textbf{Ablation Results Regarding The Objective Function for The Bounding box Regression.}}
	\footnotesize
	\setlength{\tabcolsep}{0.8mm}{
		\begin{tabular}{>{\columncolor{white}}cc|cccc}
			\toprule
			Detector & Method & mAP ($\%$) & Recall ($\%$) &  $\ell_0$ Norm &  $\ell_2$ Norm \\
			\midrule
			& \multirow{1}{*}[-0.1ex]{$\mathcal{L}_{\text{cls}}$} & 55.20  & 61.70  & \textbf{0.059} & \textbf{4.800}  \\
			& \multirow{1}{*}[-0.1ex]{$\mathcal{L}_{\text{cls}}+\mathcal{L}_{\text{CBL}}$} & 50.90  & 61.50  & \textbf{0.059} & 5.340  \\
			\rowcolor{gray!20}\cellcolor{white}\multirow{-4}{*}[-0.25ex]{FR-50} &  \multirow{1}{*}[-0.1ex]{$\mathcal{L}_{\text{cls}}+\mathcal{L}_{\text{BDL}}$} &  \textbf{47.80} & \textbf{55.80} & \textbf{0.059} & 4.990  \\
			\midrule
			& \multirow{1}{*}[-0.1ex]{$\mathcal{L}_{\text{cls}}$} & 55.80  & 70.06  & \textbf{0.059} & \textbf{5.466} \\
			& \multirow{1}{*}[-0.1ex]{$\mathcal{L}_{\text{cls}}+\mathcal{L}_{\text{CBL}}$} & 52.60  & 65.30  & \textbf{0.059} & 5.630  \\
			\rowcolor{gray!20}\cellcolor{white}\multirow{-4}{*}[-0.25ex]{FC-50} & \multirow{1}{*}[-0.1ex]{$\mathcal{L}_{\text{cls}}+\mathcal{L}_{\text{BDL}}$} & \textbf{47.00} & \textbf{61.50} & \textbf{0.059} & 5.540  \\
			\midrule
			& \multirow{1}{*}[-0.1ex]{$\mathcal{L}_{\text{cls}}$} & 42.90  & 61.90  & \textbf{0.059} & \textbf{5.010} \\
			& \multirow{1}{*}[-0.1ex]{$\mathcal{L}_{\text{cls}}+\mathcal{L}_{\text{CBL}}$} & 42.60  & 61.80  & \textbf{0.059} & 5.240  \\
			\rowcolor{gray!20}\cellcolor{white}\multirow{-4}{*}[-0.25ex]{RT-50} & \multirow{1}{*}[-0.1ex]{$\mathcal{L}_{\text{cls}}+\mathcal{L}_{\text{BDL}}$} & \textbf{39.60} & \textbf{58.00} & \textbf{0.059} & 5.170  \\
			\bottomrule
	\end{tabular}}%
	\label{tab4}%
\end{table}%
\begin{figure}[t]
	\centering
	\includegraphics[scale=.7]{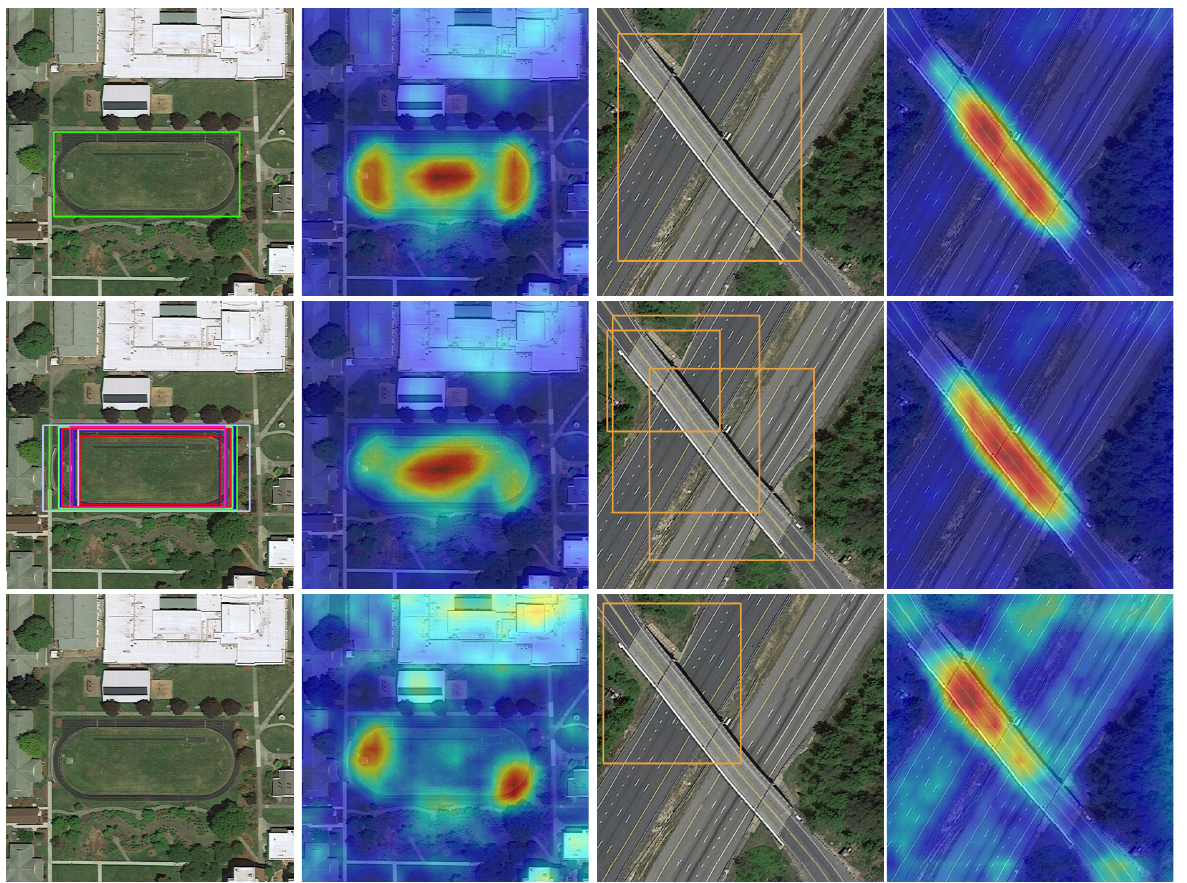}
	\caption{\textbf{Visualizations of the detected results (the Singular columns) and the corresponding features (the even columns).}~Here, the results on clean images are shown in the first row, the results on adversarial examples generated via $ \mathcal{L}_\text{cls} $ only are exhibited in the second row, and the others are the results on adversarial examples generated via $ \mathcal{L}_\text{cls} + \mathcal{L}_\text{BDL} $.}
	\label{fig8}
\end{figure}

\noindent \textbf{Ablation Study on Bbox Regression Loss.}~Here, we vary the choice of the objective function for Bbox regression to validate the importance of our proposed BDL. Specifically, we first preserve only the objective function for the classification~(denoted as $ \mathcal{L}_\text{cls} $ in~\cref{tab4}) to see the importance of attacking the Bbox regression. Then, we set another competitor,~\ie, the CBL loss~\cite{lirobust}. The results are reported in~\cref{tab4}, where two conclusions can be established. First, attacking only the classification branch shows more inferior threats than taking both Bbox regression into consideration. Meanwhile, both the introduction of BDL and CBL can sacrifice the $ \ell_2 $ norm, resulting in poorer visual quality than using $ \mathcal{L}_\text{cls} $ only. However, the improvements of the attack strengths are significant compared to the decrease of the visual quality. Thus, introducing the attack on Bbox regression poses more threats than attacking the classification only. Second, compared to CBL, we reach the significant threats with the competitive improvements of the visual quality. We can see that the $ \ell_2 $ norms of CBL are larger than our TPA, which echoes what we discussed in~\cref{sec4-3}, \ie, the problem of gradient inundation in CBL~\cite{lirobust} may risk the stagnation of optimization, leading to the decrease of the attack threats. Besides quantitative results, we visualize the effects of our BDL in~\cref{fig8}. Compared to $ \mathcal{L}_\text{cls} $ only, the addition of $ \mathcal{L}_\text{BDL} $ can result in more intense destruction, which can be reflected in the feature maps, where the attention areas are significantly interfered. By contrast, using $ \mathcal{L}_\text{cls} $ only could not cause significant attacks on intermediate representations. 
\begin{table}[tbp]
	\renewcommand\arraystretch{1.4}
	\centering
	\caption{\textbf{Ablation Results Regarding The Grid Segmentation Scheme.}}
	\footnotesize
	\setlength{\tabcolsep}{0.8mm}{
		\begin{tabular}{>{\columncolor{white}}cc|cccc}
			\toprule
			Detector & Method & mAP ($\%$) & Recall ($\%$) &  $\ell_0$ Norm &  $\ell_2$ Norm \\
			\midrule
			& $\text{U}(2)$ & 48.50  & 56.90  & 0.065  & 5.180  \\
			& $\text{U}(3)$ & 49.40  & 57.50  & \textbf{0.057} & \textbf{4.980} \\
			\rowcolor{gray!20}\cellcolor{white} & $ \text{SA}(1, 2, 3) $ & 47.80  & 55.80  & 0.059  & 4.990  \\
			\multirow{-4}{*}[-0.25ex]{FR-50} & $ \text{SA}(2, 3, 4) $ & \textbf{46.30} & \textbf{55.20} & 0.064  & 5.210  \\
			\midrule
			& $\text{U}(2)$ & 48.20  & 63.00  & 0.065  & 5.800  \\
			& $\text{U}(3)$ & 49.10  & 63.50  & \textbf{0.057} & \textbf{5.510} \\
			\rowcolor{gray!20}\cellcolor{white} & $\text{SA}(1, 2, 3)$ & 47.00  & 61.50  & 0.059  & 5.540  \\
			\multirow{-4}{*}[-0.25ex]{FC-50} & $\text{SA}(2, 3, 4)$ & \textbf{45.20} & \textbf{59.30} & 0.064  & 5.840  \\
			\midrule
			& $\text{U}(2)$ & 40.30  & 58.20  & 0.065  & 5.290  \\
			& $\text{U}(3)$ & 41.30  & 59.90  & \textbf{0.057} & \textbf{5.140} \\
			\rowcolor{gray!20}\cellcolor{white} & $\text{SA}(1, 2, 3)$ & 39.60  & 58.00  & 0.059  & 5.170  \\
			\multirow{-4}{*}[-0.25ex]{RT-50} & $\text{SA}(2, 3, 4)$ & \textbf{38.20} & \textbf{55.10} & 0.064  & 5.380  \\
			\bottomrule
	\end{tabular}}%
	\label{tab5}%
\end{table}%

\noindent \textbf{Ablation Study on Grid Segmentation Scheme.}~Since grid segmentation is the first step in patch selection scheme, and there is no researches regarding the influence of different scheme on final results. To this light, we propose to provide a preliminary experimental exploration in this subsection. Recall that we provided two options for the grid segmentation in \cref{sec4-2},~\ie, the universal scheme and the scale-adaptive scheme. In this part, we explore the influence on these schemes. Specifically, we set different parameters in $ \text{U}(n) $ and $ \text{SA}(n_1, n_2, n_3) $. The results are shown in~\cref{tab5}. Generally speaking, the choice of these schemes seems do not play a key role in the final results. When we take a closer look at these results, we can find that with the same visual effect, the scale-adaptive scheme shows more aggressive ability than the universal scheme, while leaving the visual quality get sacrificed slightly. For instance, $ \text{SA}(1, 2, 3) $ and $ \text{U}(3) $ achieve almost the same visual effect, but the attack strength of $ \text{SA}(1, 2, 3) $ is better than $ \text{U}(3) $. The same case can be found in the comparison between $ \text{U}(2) $ and $ \text{SA}(2, 3, 4) $, where $ \text{SA}(2, 3, 4) $ exhibits more threats than $ \text{U}(2) $. Thus, considering the attack strength and visual effect comprehensively, we chooses $ \text{SA}(1, 2, 3) $ as our final choice.

\section{Conclusions}
In this paper, we paid attention to PAs on object detection in O-RSIs and proposed a Threatening PA without the scarification of the visual quality, dubbed TPA. Specifically, to address the problem of inconsistency between local and global landscapes in existing patch selection schemes, we proposed to leverage the First-Order Difference of the objective function before and after masking to select the sub-patches to be attacked. Further, considering the problem of gradient inundation when applying existing coordinate-based loss to PAs directly, we designed an IoU-based objective function specific for PAs, dubbed Bounding box Drifting Loss, which pushes the detected bounding boxes far from the initial ones until there are no overlaps between them. Compared to the advanced competitor, the extensive evaluations have witnessed the remarkable effectiveness of our TPA. Moreover, we also replace the key factors of our TPA to see their influence in the final results. These comprehensive explorations also demonstrate the key role of our FOD and BDL. We hope this first attempt can arouse the research interest in further works regarding PAs on object detection in O-RSIs.

\bibliographystyle{IEEEtran}
\bibliography{ref}

\begin{thebibliography}{10}
\providecommand{\url}[1]{#1}
\csname url@samestyle\endcsname
\providecommand{\newblock}{\relax}
\providecommand{\bibinfo}[2]{#2}
\providecommand{\BIBentrySTDinterwordspacing}{\spaceskip=0pt\relax}
\providecommand{\BIBentryALTinterwordstretchfactor}{4}
\providecommand{\BIBentryALTinterwordspacing}{\spaceskip=\fontdimen2\font plus
\BIBentryALTinterwordstretchfactor\fontdimen3\font minus
  \fontdimen4\font\relax}
\providecommand{\BIBforeignlanguage}[2]{{%
\expandafter\ifx\csname l@#1\endcsname\relax
\typeout{** WARNING: IEEEtran.bst: No hyphenation pattern has been}%
\typeout{** loaded for the language `#1'. Using the pattern for}%
\typeout{** the default language instead.}%
\else
\language=\csname l@#1\endcsname
\fi
#2}}
\providecommand{\BIBdecl}{\relax}
\BIBdecl

\bibitem{szegedy2016rethinking}
C.~Szegedy, V.~Vanhoucke, S.~Ioffe, J.~Shlens, and Z.~Wojna, ``Rethinking the
  inception architecture for computer vision,'' in \emph{Proc. IEEE Conf.
  Comput. Vis. Pattern Recognit.}, 2016, pp. 2818--2826.

\bibitem{szegedy2017inception}
C.~Szegedy, S.~Ioffe, V.~Vanhoucke, and A.~A. Alemi, ``Inception-v4,
  inception-resnet and the impact of residual connections on learning,'' in
  \emph{AAAI}, 2017.

\bibitem{ren2015faster}
S.~Ren, K.~He, R.~Girshick, and J.~Sun, ``Faster r-cnn: Towards real-time
  object detection with region proposal networks,'' in \emph{Proc. Adv. Neural
  Inform. Process. Syst.}, vol.~28, 2015.

\bibitem{tian2019fcos}
Z.~Tian, C.~Shen, H.~Chen, and T.~He, ``Fcos: Fully convolutional one-stage
  object detection,'' in \emph{Proc. Int. Conf. Comput. Vis.}, 2019, pp.
  9627--9636.

\bibitem{lin2017focal}
T.-Y. Lin, P.~Goyal, R.~Girshick, K.~He, and P.~Doll{\'a}r, ``Focal loss for
  dense object detection,'' in \emph{Proc. Int. Conf. Comput. Vis.}, 2017, pp.
  2980--2988.

\bibitem{bochkovskiy2020yolov4}
A.~Bochkovskiy, C.-Y. Wang, and H.-Y.~M. Liao, ``Yolov4: Optimal speed and
  accuracy of object detection,'' \emph{arXiv:2004.10934}, 2020.

\bibitem{redmon2016you}
J.~Redmon, S.~Divvala, R.~Girshick, and A.~Farhadi, ``You only look once:
  Unified, real-time object detection,'' in \emph{Proc. IEEE Conf. Comput. Vis.
  Pattern Recognit.}, 2016, pp. 779--788.

\bibitem{he2016deep}
K.~He, X.~Zhang, S.~Ren, and J.~Sun, ``Deep residual learning for image
  recognition,'' in \emph{Proc. IEEE Conf. Comput. Vis. Pattern Recognit.},
  Jun. 2016, pp. 770--778.

\bibitem{xie2017aggregated}
S.~Xie, R.~Girshick, P.~Doll{\'a}r, Z.~Tu, and K.~He, ``Aggregated residual
  transformations for deep neural networks,'' in \emph{Proc. IEEE Conf. Comput.
  Vis. Pattern Recognit.}, Jul. 2017, pp. 1492--1500.

\bibitem{cheng2022class}
G.~Cheng, P.~Lai, D.~Gao, and J.~Han, ``Class attention network for image
  recognition,'' \emph{Sci. China Inf. Sci.}, 2022.

\bibitem{cheng2021anchor}
G.~Cheng, J.~Wang, K.~Li, X.~Xie, C.~Lang, Y.~Yao, and J.~Han, ``Anchor-free
  oriented proposal generator for object detection,'' \emph{IEEE Trans. Geosci.
  Remote Sens.}, 2021.

\bibitem{xia2018dota}
G.-S. Xia, X.~Bai, J.~Ding, Z.~Zhu, S.~Belongie, J.~Luo, M.~Datcu, M.~Pelillo,
  and L.~Zhang, ``Dota: A large-scale dataset for object detection in aerial
  images,'' in \emph{Proceedings of the IEEE conference on computer vision and
  pattern recognition}, 2018, pp. 3974--3983.

\bibitem{li2022semi}
J.~Li, Y.~Liao, J.~Zhang, D.~Zeng, and X.~Qian, ``Semi-supervised degan for
  optical high-resolution remote sensing image scene classification,''
  \emph{Remote Sens.}, vol.~14, no.~17, p. 4418, 2022.

\bibitem{niu2022multi}
B.~Niu, Z.~Pan, J.~Wu, Y.~Hu, and B.~Lei, ``Multi-representation dynamic
  adaptation network for remote sensing scene classification,'' \emph{IEEE
  Trans. Geosci. Remote Sens.}, vol.~60, pp. 1--19, 2022.

\bibitem{pei2021multi}
L.~Pei, G.~Cheng, X.~Sun, Q.~Li, M.~Zhang, and S.~Miao, ``Multi-scale
  bidirectional feature fusion for one-stage oriented object detection in
  aerial images,'' in \emph{IGARSS}, 2021, pp. 2592--2595.

\bibitem{li2023instance}
C.~Li, G.~Cheng, G.~Wang, P.~Zhou, and J.~Han, ``Instance-aware distillation
  for efficient object detection in remote sensing images,'' \emph{IEEE Trans.
  Geosci. Remote Sens.}, 2023.

\bibitem{yao2022improving}
Y.~Yao, G.~Cheng, G.~Wang, S.~Li, P.~Zhou, X.~Xie, and J.~Han, ``On improving
  bounding box representations for oriented object detection,'' \emph{IEEE
  Trans. Geosci. Remote Sens.}, 2022.

\bibitem{goodfellow2014explaining}
I.~J. Goodfellow, J.~Shlens, and C.~Szegedy, ``Explaining and harnessing
  adversarial examples,'' \emph{arXiv:1412.6572}, 2014.

\bibitem{szegedy2013intriguing}
C.~Szegedy, W.~Zaremba, I.~Sutskever, J.~Bruna, D.~Erhan, I.~Goodfellow, and
  R.~Fergus, ``Intriguing properties of neural networks,''
  \emph{arXiv:1312.6199}, 2013.

\bibitem{carlini2017towards}
N.~Carlini and D.~Wagner, ``Towards evaluating the robustness of neural
  networks,'' in \emph{IEEE Symp. Security Privacy}.\hskip 1em plus 0.5em minus
  0.4em\relax IEEE, 2017, pp. 39--57.

\bibitem{madry2017towards}
A.~Madry, A.~Makelov, L.~Schmidt, D.~Tsipras, and A.~Vladu, ``Towards deep
  learning models resistant to adversarial attacks,'' in \emph{Proc. Int. Conf.
  Learn. Represent.}, 2018.

\bibitem{moosavi2016deepfool}
S.-M. Moosavi-Dezfooli, A.~Fawzi, and P.~Frossard, ``Deepfool: a simple and
  accurate method to fool deep neural networks,'' in \emph{Proc. IEEE Conf.
  Comput. Vis. Pattern Recognit.}, 2016, pp. 2574--2582.

\bibitem{tramer2018ensemble}
F.~Tram{\`e}r, A.~Kurakin, N.~Papernot, I.~Goodfellow, D.~Boneh, and
  P.~McDaniel, ``Ensemble adversarial training: Attacks and defenses,'' in
  \emph{Proc. Int. Conf. Learn. Represent.}, 2018.

\bibitem{cheng2021perturbation}
G.~Cheng, X.~Sun, K.~Li, L.~Guo, and J.~Han, ``Perturbation-seeking generative
  adversarial networks: A defense framework for remote sensing image scene
  classification,'' \emph{IEEE Trans. Geosci. Remote Sens.}, vol.~60, pp.
  1--11, 2021.

\bibitem{xu2022ai}
Y.~Xu, T.~Bai, W.~Yu, S.~Chang, P.~M. Atkinson, and P.~Ghamisi, ``Ai security
  for geoscience and remote sensing: Challenges and future trends,''
  \emph{arXiv:2212.09360}, 2022.

\bibitem{czaja2018adversarial}
W.~Czaja, N.~Fendley, M.~Pekala, C.~Ratto, and I.-J. Wang, ``Adversarial
  examples in remote sensing,'' in \emph{Proc. SIGSPATIAL Int. Conf. Adv.
  Geogr. Inf. Syst.}, 2018, pp. 408--411.

\bibitem{xu2020assessing}
Y.~Xu, B.~Du, and L.~Zhang, ``Assessing the threat of adversarial examples on
  deep neural networks for remote sensing scene classification: Attacks and
  defenses,'' \emph{IEEE Trans. Geosci. Remote Sens.}, vol.~59, no.~2, pp.
  1604--1617, 2020.

\bibitem{xie2017adversarial}
C.~Xie, J.~Wang, Z.~Zhang, Y.~Zhou, L.~Xie, and A.~Yuille, ``Adversarial
  examples for semantic segmentation and object detection,'' in
  \emph{Proceedings of the IEEE international conference on computer vision},
  2017, pp. 1369--1378.

\bibitem{chen2019shapeshifter}
S.-T. Chen, C.~Cornelius, J.~Martin, and D.~H.~P. Chau, ``Shapeshifter: Robust
  physical adversarial attack on faster r-cnn object detector,'' in \emph{Proc.
  ECML PKDD}.\hskip 1em plus 0.5em minus 0.4em\relax Springer, 2019, pp.
  52--68.

\bibitem{lirobust}
Y.~Li, D.~Tian, X.~Bian, and S.~Lyu, ``Robust adversarial perturbation on deep
  proposal-based models,'' in \emph{Proc. Brit. Mach. Vis. Conf.}

\bibitem{zhang2020contextual}
H.~Zhang, W.~Zhou, and H.~Li, ``Contextual adversarial attacks for object
  detection.''\hskip 1em plus 0.5em minus 0.4em\relax IEEE, 2020, pp. 1--6.

\bibitem{nezami2021pick}
O.~M. Nezami, A.~Chaturvedi, M.~Dras, and U.~Garain, ``Pick-object-attack:
  Type-specific adversarial attack for object detection,'' \emph{Comput. Vis.
  Image Underst.}, vol. 211, p. 103257, 2021.

\bibitem{wu2020dpattack}
S.~Wu, T.~Dai, and S.-T. Xia, ``Dpattack: Diffused patch attacks against
  universal object detection,'' \emph{arXiv:2010.11679}, 2020.

\bibitem{zhao2020object}
Y.~Zhao, H.~Yan, and X.~Wei, ``Object hider: Adversarial patch attack against
  object detectors,'' \emph{arXiv:2010.14974}, 2020.

\bibitem{huang2021rpattack}
H.~Huang, Y.~Wang, Z.~Chen, Z.~Tang, W.~Zhang, and K.-K. Ma, ``Rpattack:
  Refined patch attack on general object detectors,'' 2021, pp. 1--6.

\bibitem{brown2017adversarial}
T.~B. Brown, D.~Man{\'e}, A.~Roy, M.~Abadi, and J.~Gilmer, ``Adversarial
  patch,'' \emph{arXiv:1712.09665}, 2017.

\bibitem{liu2018dpatch}
X.~Liu, H.~Yang, Z.~Liu, L.~Song, H.~Li, and Y.~Chen, ``Dpatch: An adversarial
  patch attack on object detectors,'' \emph{arXiv:1806.02299}, 2018.

\bibitem{lee2019physical}
M.~Lee and Z.~Kolter, ``On physical adversarial patches for object detection,''
  \emph{arXiv:1906.11897}, 2019.

\bibitem{lin2017feature}
T.-Y. Lin, P.~Doll{\'a}r, R.~Girshick, K.~He, B.~Hariharan, and S.~Belongie,
  ``Feature pyramid networks for object detection,'' in \emph{Proceedings of
  the IEEE conference on computer vision and pattern recognition}, 2017, pp.
  2117--2125.

\bibitem{dong2019evading}
Y.~Dong, T.~Pang, H.~Su, and J.~Zhu, ``Evading defenses to transferable
  adversarial examples by translation-invariant attacks,'' in \emph{Proc. IEEE
  Conf. Comput. Vis. Pattern Recognit.}, 2019, pp. 4312--4321.

\bibitem{xie2019improving}
C.~Xie, Z.~Zhang, Y.~Zhou, S.~Bai, J.~Wang, Z.~Ren, and A.~L. Yuille,
  ``Improving transferability of adversarial examples with input diversity,''
  in \emph{Proc. IEEE Conf. Comput. Vis. Pattern Recognit.}, 2019, pp.
  2730--2739.

\bibitem{lin2019nesterov}
J.~Lin, C.~Song, K.~He, L.~Wang, and J.~E. Hopcroft, ``Nesterov accelerated
  gradient and scale invariance for adversarial attacks,'' in \emph{Proc. Int.
  Conf. Learn. Represent.}, 2020.

\bibitem{dong2018boosting}
Y.~Dong, F.~Liao, T.~Pang, H.~Su, J.~Zhu, X.~Hu, and J.~Li, ``Boosting
  adversarial attacks with momentum,'' in \emph{Proc. IEEE Conf. Comput. Vis.
  Pattern Recognit.}, 2018, pp. 9185--9193.

\bibitem{chen2020hopskipjumpattack}
J.~Chen, M.~I. Jordan, and M.~J. Wainwright, ``Hopskipjumpattack: A
  query-efficient decision-based attack,'' in \emph{Proc. IEEE Symp. Secur.
  Priv.}\hskip 1em plus 0.5em minus 0.4em\relax IEEE, 2020, pp. 1277--1294.

\bibitem{brendel2017decision}
W.~Brendel, J.~Rauber, and M.~Bethge, ``Decision-based adversarial attacks:
  Reliable attacks against black-box machine learning models,''
  \emph{arXiv:1712.04248}, 2017.

\bibitem{SUN2022108728}
X.~Sun, G.~Cheng, L.~Pei, and J.~Han, ``Query-efficient decision-based attack
  via sampling distribution reshaping,'' \emph{Pattern Recognit.}, p. 108728,
  2022.

\bibitem{brunner2019guessing}
T.~Brunner, F.~Diehl, M.~T. Le, and A.~Knoll, ``Guessing smart: Biased sampling
  for efficient black-box adversarial attacks,'' in \emph{Int. Conf. Comput.
  Vis.}, 2019, pp. 4958--4966.

\bibitem{sun2022exploring}
X.~Sun, G.~Cheng, H.~Li, L.~Pei, and J.~Han, ``Exploring effective data for
  surrogate training towards black-box attack,'' in \emph{Proc. IEEE Conf.
  Comput. Vis. Pattern Recognit.}, 2022, pp. 15\,355--15\,364.

\bibitem{zhou2020dast}
M.~Zhou, J.~Wu, Y.~Liu, S.~Liu, and C.~Zhu, ``Dast: Data-free substitute
  training for adversarial attacks,'' in \emph{Proc. IEEE Conf. Comput. Vis.
  Pattern Recognit.}, 2020, pp. 234--243.

\bibitem{silva2020opportunities}
S.~H. Silva and P.~Najafirad, ``Opportunities and challenges in deep learning
  adversarial robustness: A survey,'' \emph{arXiv:2007.00753}, 2020.

\bibitem{athalye2018synthesizing}
A.~Athalye, L.~Engstrom, A.~Ilyas, and K.~Kwok, ``Synthesizing robust
  adversarial examples,'' in \emph{Proc. Int. Conf. Mach. Learn.}\hskip 1em
  plus 0.5em minus 0.4em\relax PMLR, 2018, pp. 284--293.

\bibitem{liu2019perceptual}
A.~Liu, X.~Liu, J.~Fan, Y.~Ma, A.~Zhang, H.~Xie, and D.~Tao,
  ``Perceptual-sensitive gan for generating adversarial patches,'' in
  \emph{AAAI}, vol.~33, no.~01, 2019, pp. 1028--1035.

\bibitem{lian2022benchmarking}
J.~Lian, S.~Mei, S.~Zhang, and M.~Ma, ``Benchmarking adversarial patch against
  aerial detection,'' \emph{IEEE Trans. Geosci. Remote Sens.}, vol.~60, pp.
  1--16, 2022.

\bibitem{den2020adversarial}
R.~den Hollander, A.~Adhikari, I.~Tolios, M.~van Bekkum, A.~Bal, S.~Hendriks,
  M.~Kruithof, D.~Gross, N.~Jansen, G.~Perez \emph{et~al.}, ``Adversarial patch
  camouflage against aerial detection,'' in \emph{Artificial Intelligence and
  Machine Learning in Defense Applications II}, vol. 11543.\hskip 1em plus
  0.5em minus 0.4em\relax SPIE, 2020, pp. 77--86.

\bibitem{lu2021scale}
M.~Lu, Q.~Li, L.~Chen, and H.~Li, ``Scale-adaptive adversarial patch attack for
  remote sensing image aircraft detection,'' \emph{Remote Sens.}, vol.~13,
  no.~20, p. 4078, 2021.

\bibitem{thys2019fooling}
S.~Thys, W.~Van~Ranst, and T.~Goedem{\'e}, ``Fooling automated surveillance
  cameras: adversarial patches to attack person detection,'' in \emph{Proc.
  IEEE Conf. Comput. Vis. Pattern Recognit. Workshops}, 2019, pp. 0--0.

\bibitem{wang2021dual}
J.~Wang, A.~Liu, Z.~Yin, S.~Liu, S.~Tang, and X.~Liu, ``Dual attention
  suppression attack: Generate adversarial camouflage in physical world,'' in
  \emph{Proc. IEEE Conf. Comput. Vis. Pattern Recognit.}, 2021, pp. 8565--8574.

\bibitem{kurakin2018adversarial}
A.~Kurakin, I.~J. Goodfellow, and S.~Bengio, ``Adversarial examples in the
  physical world,'' in \emph{Artif. Intell. Safety Secur.}\hskip 1em plus 0.5em
  minus 0.4em\relax Chapman and Hall/CRC, 2018, pp. 99--112.

\bibitem{lin2014microsoft}
T.-Y. Lin, M.~Maire, S.~Belongie, J.~Hays, P.~Perona, D.~Ramanan,
  P.~Doll{\'a}r, and C.~L. Zitnick, ``Microsoft coco: Common objects in
  context,'' in \emph{Proc. Eur. Conf. Comput. Vis.}\hskip 1em plus 0.5em minus
  0.4em\relax Springer, 2014, pp. 740--755.

\end{thebibliography}
\end{document}